\pdfoutput=1

\documentclass[11pt]{article}

\usepackage[]{acl}

\usepackage{times}
\usepackage{latexsym}

\usepackage[T1]{fontenc}
\usepackage{graphicx}

\usepackage[utf8]{inputenc}

\usepackage{microtype}

\title{BERT-VQA: Visual Question Answering on Plots}

\author{Tai Vu \\
  Stanford University \\
  \texttt{taivu@stanford.edu} \\\And
  Robert Yang \\
  Stanford University  \\  
  \texttt{bobyang9@stanford.edu} 
}

\begin{document}
\maketitle
\begin{abstract}
Visual question answering has been an exciting challenge in the field of natural language understanding, as it requires deep learning models to exchange information from both vision and language domains. In this project, we aim to tackle a subtask of this problem, namely visual question answering on plots. To achieve this, we developed \textbf{BERT-VQA}, a VisualBERT-based model architecture with a pretrained ResNet 101 image encoder, along with a potential addition of joint fusion. We trained and evaluated this model against a baseline that consisted of a LSTM, a CNN, and a shallow classifier. The final outcome disproved our core hypothesis that the cross-modality module in VisualBERT is essential in aligning plot components with question phrases. Therefore, our work provided valuable insights into the difficulty of the plot question answering challenge as well as the appropriateness of different model architectures in solving this problem.
\end{abstract}

\section{Introduction}

Visual question answering (VQA) is a multimodal task and area of research. The aim is for the machine to answer text-based questions (usually approached by NLU) informed by a related image (usually processed with computer vision). This is a fascinating area of research over the last few years, as it require sophisticated model architectures that can perform joint inference over language and vision data. We are especially interested in visual question answering on plots, which is illustrated in \citet{plotqa}. 

Visual question answering on plots is useful in terms of basic scientific value as it is a very difficult task that requires some capability of reasoning. An important example is associating a key label to parts of the graph. To do so, a human must first notice that the key label and the key object (usually a line, dot, etc.) are related, and further associate the key object with similar objects on the plot. Finally, it can then associate the label itself with the related plot points or items.

This problem seems even more important in practical applications such as search (reading diagrams among the search results to determine webpage relevancy) and having a better machine understanding of scientific literature, but these complex applications are outside the scope of our $4$-week project.

In this paper, we aim to tackle the task of visual question answering on plots using deep neural networks by designing a BERT-like machine learning system called \textbf{BERT-VQA}. In visual question answering, we have two inputs, including the question text and the related image.  We also have one output: the answer to the question text conditioned on the related image. Plot question answering is a subclass of visual question answering, where the related image must be a plot. 

The central hypothesis for this paper is that VisualBERT, a state-of-the-art model for cross-modality, will be highly performant on question answering for plots. The rationale for this hypothesis is that cross-modality is very important for plot question-answering, with the questions usually requiring logical understanding of the relationships of plot elements and parts of the text question. Historically, this cross-modality was achieved by fusing visual and text embeddings together, and later by using an attention mechanism. We use the VisualBERT model, which is a development upon attention mechanism which we guess can improve the cross-modality inference. Especially important is the property of VisualBERT to associate visual and text elements early on, allowing more meaningful logical links to form later on. We think this property would be critical for plot visual question-answering since there can be many layers of logic in one question (for example, relating the question to the key of the plot, to the individual plot elements like bars or lines, to the axis label text, to how these text elements correspond to the answer of the question). Gaining knowledge regarding this hypothesis is important because it sheds light on the role of BERT and other modern mechanisms used for multi-modal fusion - whether a more advanced and general mechanism for fusion is beneficial, or if something more primitive, like concatenation and subsequent processing in a multi-layer perceptron is sufficient. The conclusions in this paper, although limited in scope to the domain of plot question-answering, can provide suggestions for the greater direction of producing effective yet simple models for multi-modal tasks.

We take a few steps to address this hypothesis. We have implemented both the traditional perspective of fusion (concatenation) and the modern perspective of fusion (VisualBERT). We trained both types of models and compared them on plot question-answering. The results will show if multiple Transformer layers as a fusion mechanism is necessary, unnecessary, or detrimental, for multi-modal fusion of plot data and the text of questions. We also took steps to make these results as fair as possible, including keeping the architectures as similar as possible outside of the fusion location. This means using the same type of ResNet, ResNet 101, with the same pretrained parameters, and using the same architecture for the shallow classifier at the end (barring minor implementation details such as batch size, dropout, and the learning rate hyperparameters optimized for each model type). 

We found that using BERT as a fusion mechanism through the VisualBERT architecture was actually detrimental to the performance of the system, using many times the resources but scoring 2.96 percentage points lower in accuracy, 4.00 percentage points lower in precision, 2.66 percentage points lower in recall, and 2.82 percent points lower in F1 score than the traditional concatenation fusion method we used as the baseline. This means that the VisualBERT fusion mechanism is not as effective as concatenation in allowing associations to be made among the visual and text data. Hence, our core hypothesis was shown to be incorrect.

\section{Prior Literature}

The main challenge in visual question answering is how to combine knowledge from both vision and language domains. A popular solution is using a cross-modality network that aggregates information and learns multimodal interactions. For instance, \citet{li2019visualbert} proposes VisualBERT, a BERT-like model architecture that jointly extracts information from visual and textual inputs. In particular, the model adds a pre-trained object localization network like Faster-RCNN to produce visual feature representations of bounding boxes in the input images. Subsequently, these bounding region encodings are combined with the set of textual tokens and then fed into a stack of Transformer layers from a BERT network. These layers will then output joint contextual representations of both image and text data, which can be utilized in downstream tasks like visual question answering or image captioning. Similarly, \citet{tan2019lxmert} develops a vision-and-language cross-modality framework called LXMERT. This systems leverages a the cross-modality encoder with  stack of self-attention and bi-directional cross-attention sub-layers for learning to exchange and align the information from the above vision and language representations. Both of the aforementioned models are pretrained on massive datasets like the COCO image caption database \citep{chen2015microsoft} and then fine-tuned on downstream tasks, which allows them learn compact vector representations that captures highly complex vision-and-language connections. Other similar approaches include UNITER \citep{chen2020uniter} and MUTAN \citep{ben2017mutan}.

In addition, a novel and fascinating idea that is the application of graph neural networks (GNN). For example, \citet{li2019relation} suggests ReGAT, a visual question answering system that has ability to capture multi-type inter-object relationships in visual scenes with a clever application of graph neural networks. The novel and most important component of the model is its relation encoder, which produces region-level, relation-aware, question-adaptive representations from the visual inputs. In fact, the relation encoder represents image data as graphs and leverages graph attention networks (GAT) to learn both explicit and implicit relations amongst the detected bounding boxes. In this case, explicit relations are geometric positions and semantic interactions, while implicit ones represent hidden dynamics amongst objects. Furthermore, ReGAT guarantees that all the learned relations are question-adaptive by injecting semantic information from the questions into the relation encodings. Hence, the feature representations absorb both visual contents about objects and semantic clues in the questions, so only object relations that are most relevant to input questions are captured. In this way, even though the input data do not contain a graph structure, ReGAT successfully models dynamics between visual objects as a graph and utilizes graph attention networks  to gain valuable insights about single-modality and cross-modality dynamics. Similar insights have been explored in recent studies, such as \citet{teney2017graph} and \citet{narasimhan2018out}. This trend demonstrates that GNNs have potentials to strengthen the models' ability to fully understand visual scenes by extracting inter-object relationships beyond static object localization.




Although there has been extensive machine learning research in visual question answering in general, the task of visual question answering on plots still remains a relatively unexplored challenge, because the state of the art is only around $1/4$ of the human accuracy. In particular, \citet{plotqa} provides a model to approach the task of question answering on the new PlotQA dataset. The paper recognizes the large range of difficulty of the plot visual question answering task and uses a multi-stage model to increase performance. First, a binary classifier is trained to decide between using a simple or complex model. The simple model is designed for easier questions that don’t involve multiple levels of reasoning. It consists of the older-style setup of a convolutional neural network to process the image and LSTM to process the question, then combining the results and adding a few more layers at the end to produce the right answer. The complex model recognizes that a lot of reasoning has to do with text baked into the image, so it includes a semi-hierarchical approach. The image is first passed through a “visual elements” detector (anything in the plot such as title, labels, key, lines, points, etc.); this module is a fine-tuned object detection model (the paper used Faster R-CNN + FPN). Next, it is passed through OCR to find the text baked into the plot. Afterwards, this text and the objects detected in the first stage is used to do reasoning on which objects are linked (they call this the SIE). Then, they put all the information into a table; this reduces the problem to table question answering which is a standard problem. This paper is highly complex and they train a lot of parts where each one has to work well for all of them to work well; however, it does seem to have higher accuracy compared to the previous works (though accuracy still remains very low, at about $22.52\%$). Hence, we concluded from all these papers that a cleaner method of multi-modal fusion is needed, with as less feature engineering as possible, and that may be possible with VisualBERT.

\section{Data}
Our dataset is a subset of the PlotQA dataset by \citet{plotqa}. PlotQA is a dataset targeted towards complex reasoning with visual question answering specifically on plots \citet{plotqa}. It includes a diverse arrangement of plots, including bar plots, line plots, and dot-line plots. 

\begin{figure}[h]
\centering
\includegraphics[scale=0.15]{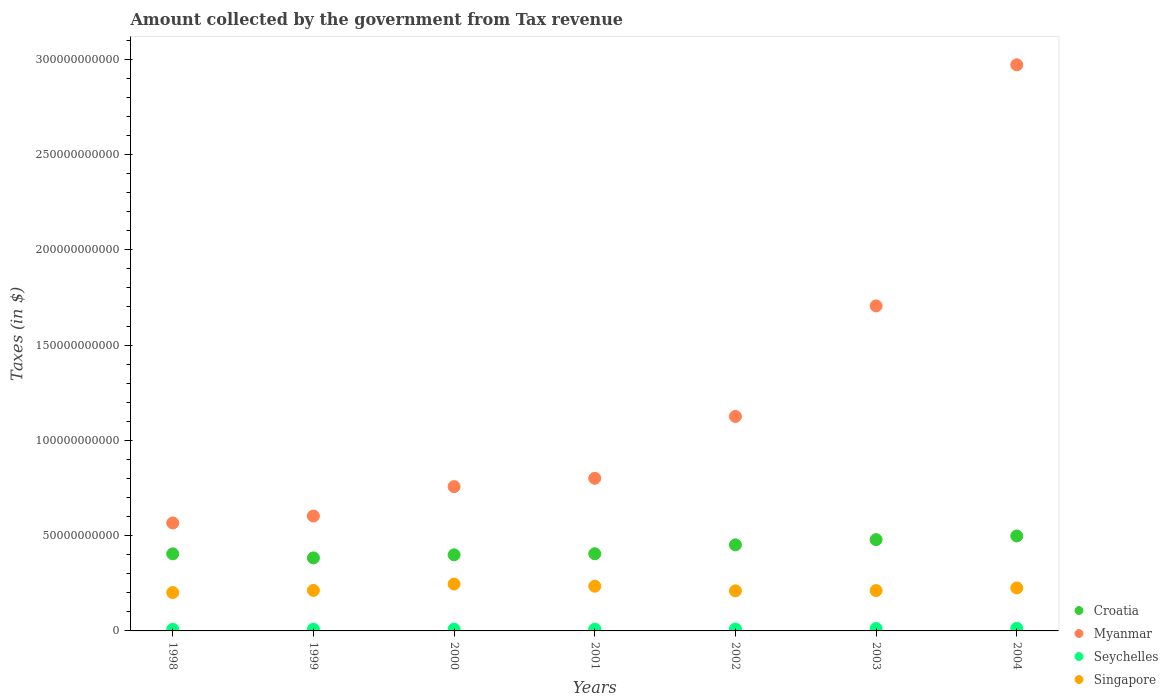}
\caption{Example dot-line plot.}
\label{fig:plot1}
\end{figure}

\begin{figure}[h]
\centering
\includegraphics[scale=0.15]{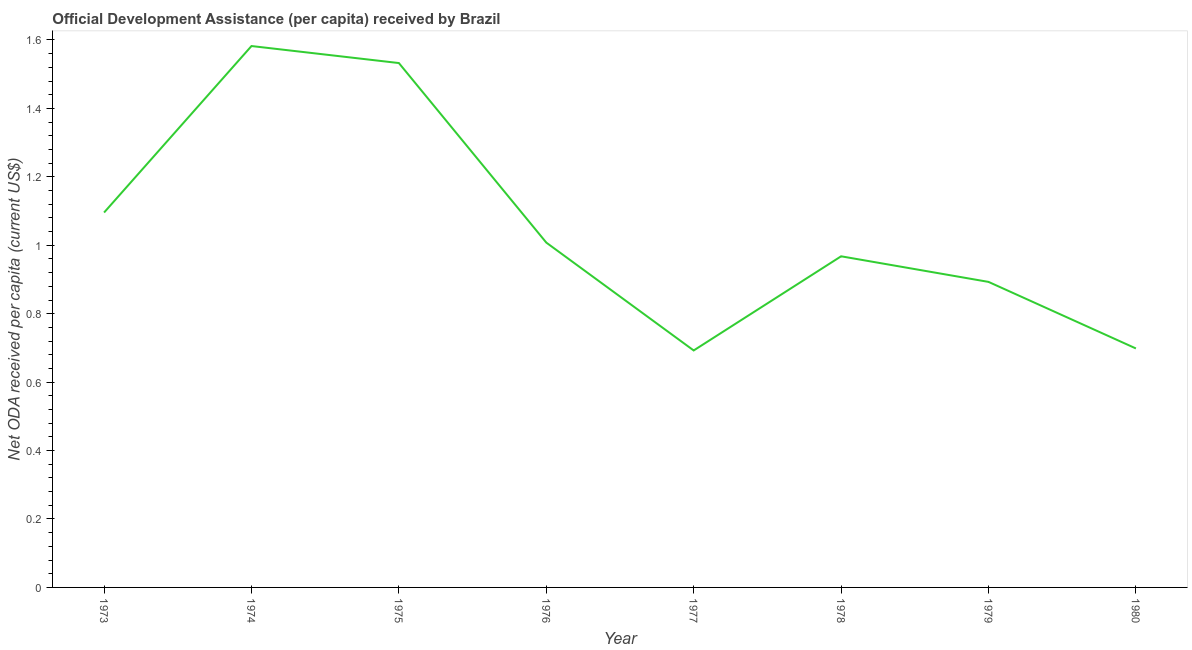}
\caption{Example line plot.}
\label{fig:plot2}
\end{figure}

\begin{figure}[h]
\centering
\includegraphics[scale=0.15]{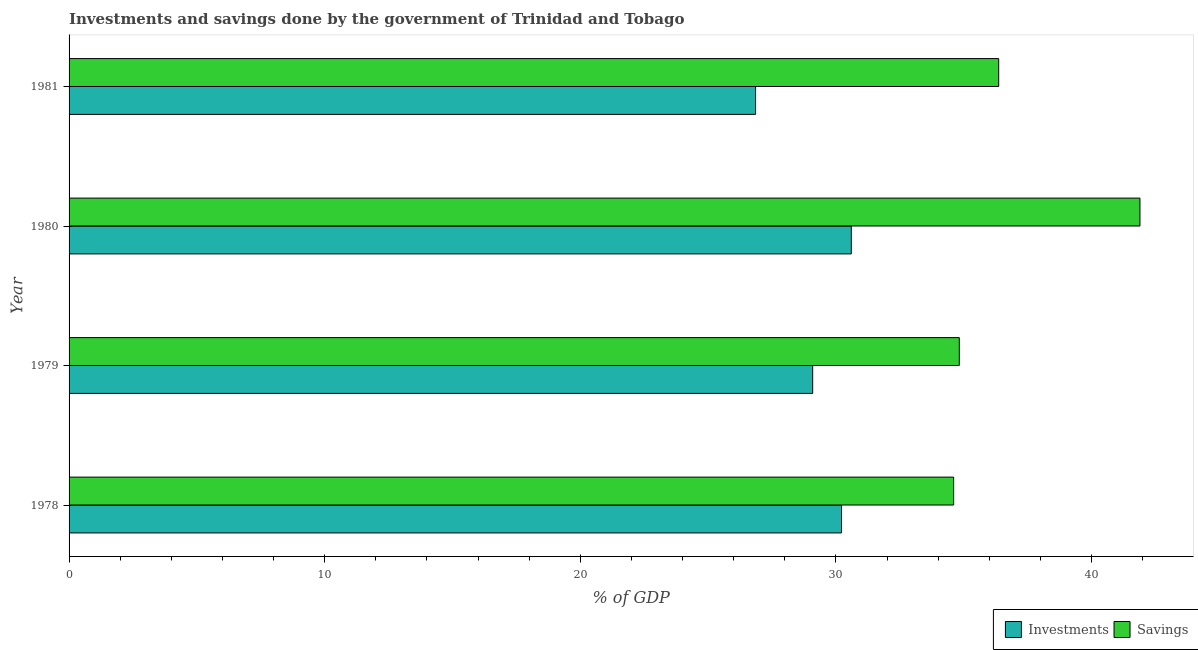}
\caption{Example bar plot.}
\label{fig:plot3}
\end{figure}

It includes questions that can have one of three answer types: “Yes/No”, “fixed vocabulary”, and “open vocabulary”. For this project, we are going to use only the “Yes/No” questions to test our hypothesis. There are $1,119,713$ “Yes/No” questions in the dataset, of which $784,115$ are in the training set, $167,871$ are in the validation set, and $167,727$ are in the test set. Due to computational constraints, our final dataset is a subset of these 1119713 questions. Our training set contains $87,962$ questions ($43,981$ labeled "Yes" and $43,981$ labeled "No") on $10,000$ images, our validation set contains $43,770$ questions ($21,885$  labeled "Yes" and $21,885$ labeled "No") on $5,000$ images, and our test set contains $43,918$ questions ($21,959$ labeled "Yes" and $21,959$ labeled "No") on $5,000$ images. 

PlotQA aims to create a challenge for the model through three different types of questions:  “structure”, “data retrieval”, and “reasoning”. “Structure” questions test if the model can associate the relative positions of graph elements with the question prompt. “Data Retrieval” questions test if the model can extract information from the graph in a direct manner. “Reasoning” tests if the model can identify the semantics of various areas of the graph and apply logic to these semantics. Of the $87,962$ “Yes/No” questions in the finalized training set, $37,644$ are “Structure” type questions, $14,476$ are “Data Retrieval” questions”, and $35,842$ are “Reasoning” type questions. Of the $43,770$ “Yes/No” questions in the finalized validation set, $18,628$ are “Structure” type questions, $7,242$ are “Data Retrieval” questions”, and $17,900$ are “Reasoning” type questions. Of the $43,918$ “Yes/No” questions in the finalized test set, $18,814$ are “Structure” type questions, $7,219$ are “Data Retrieval” questions”, and $17,885$ are “Reasoning” type questions. 

\begin{figure}[h]
\centering
\includegraphics[scale=0.15]{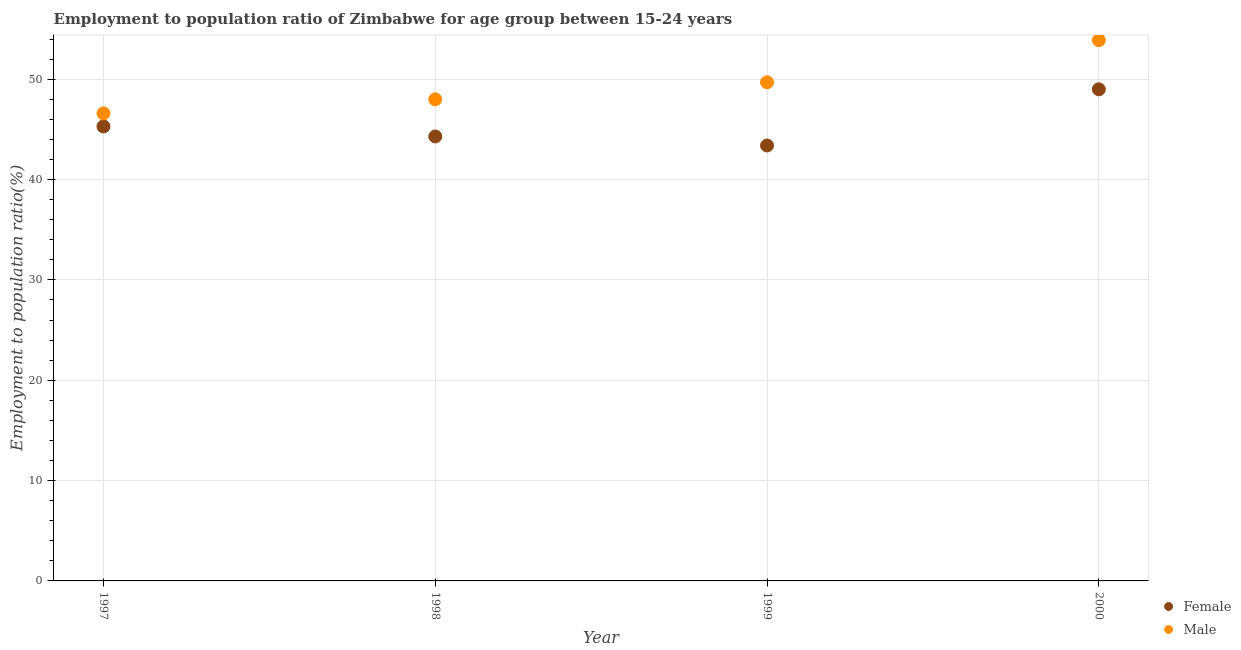}
\caption{Example plot with all three different types of questions.}
\label{fig:plot4}
\end{figure}

As an example, relative to \ref{fig:plot4}, the question "Is the number of dotlines equal to the number of legend labels?" is a "Structure" question. The question "Is the employment to population ratio (male) in 1997 less than that in 1999?" is a "Data Retrieval" question. Finally, the question "Does the employment to population ratio(female) monotonically increase over the years?" is a "Reasoning" (also known as "Compound") question.

\section{Models}

\subsection{Baseline Model: LSTM + CNN + Shallow Classifier}

\begin{figure}[h]
\centering
\includegraphics[scale=0.17]{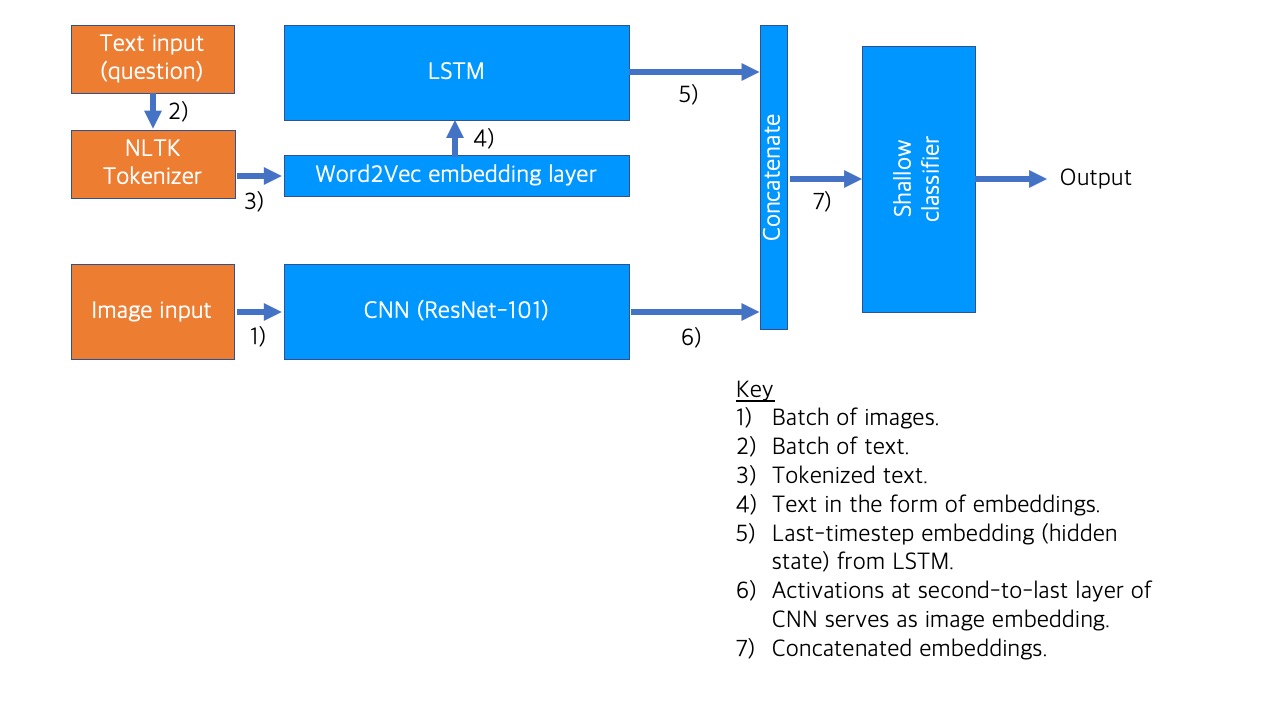}
\caption{Model architecture of the baseline model (LSTM + CNN + Shallow Classifier)}
\label{fig:plot5}
\end{figure}

LSTM + CNN + Shallow Classifier is one of the most popular models for multimodal tasks, including visual question-answering. We consulted multiple sources, including the general framework introduced in \citet{teney2018tips} and the “easy stream” model introduced in PlotQA \citet{plotqa}. We then simplified the model further to include only three components: LSTM for obtaining the embeddings of the question, CNN for obtaining one embedding summarizing the entire image, and a shallow classifier at the end whose input is the concatenation of the embedding vectors from the LSTM and CNN. As the type and location of the attention mechanism in the model varies from architecture to architecture in the models we consulted, we decided to remove attention entirely, which makes this baseline more similar to the earliest deep models in visual question-answering. TODO: source needed

The diagram in Figure \ref{fig:plot5} shows the concrete layout and architecture of the LSTM + CNN + Classifier baseline. In step 1, the image input (consisting of the plot image) is fed into the ResNet 101 CNN. Simultaneously, in step 2 and step 3, the text input is tokenized and converted into GloVe embeddings \cite{pennington2014glove}. In step 4, the embeddings are sent into the LSTM, from which the hidden state at the final timestep is taken and concatenated with the image embedding from the ResNet 101 CNN. Finally, a shallow network is applied to the concatenation which yields a binary prediction (corresponding to “Yes”/”No”). 




\subsection{Proposed Model: Modified VisualBERT}

\begin{figure}[h]
\centering
\includegraphics[scale=0.15]{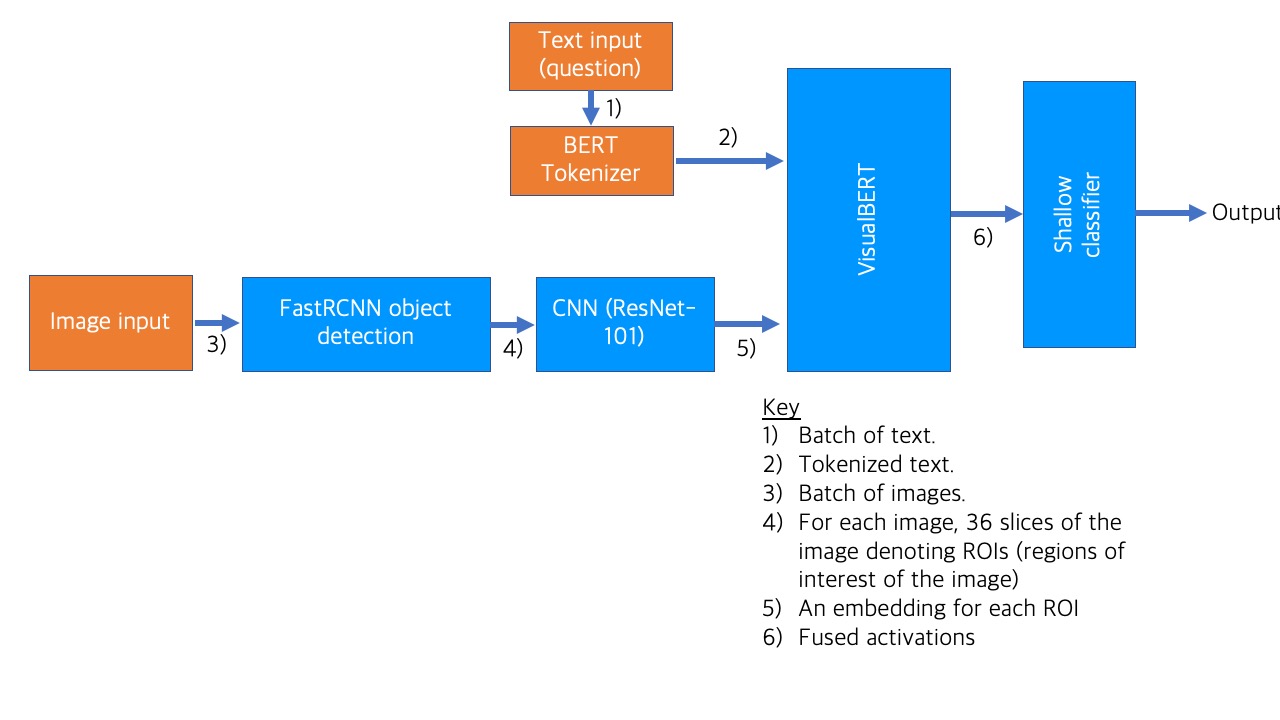}
\caption{Model architecture of the modified VisualBERT}
\label{fig:plot6}
\end{figure}

This is the first of two models which we propose will increase the accuracy of PlotQA. Given the baseline model, we formed a hypothesis that more sophisticated visual and textual encoding models would allow for better understanding of the inputs. We also believed that a cross-modality module, which contained self-attention and cross-attention layers, was helpful in exchanging and aligning the information from the vision and language representations. Therefore, we selected a combination of Faster R-CNN, ResNet 101, and VisualBERT for these encoders. 

In particular, we implemented a model architecture that follows the VisualBERT architecture \citep{li2019visualbert}. Our model used a pretrained Faster R-CNN to extract bounding boxes, which are fed into a pre-trained ResNet to obtain visual embeddings \citep{ren2015faster, he2016deep}. However, instead of using ResNet 152 as suggested by \citet{li2019visualbert}, we decided to utilize a ResNet 101, which is a part of a pretrained LXMERT model \citep{tan2019lxmert}. The reason is that LXMERT has been pretrained on a large set of image-and-sentence pairs from MS COCO  and Visual Genome captioning databases using several novel pretraining objectives \citep{lin2014microsoft, krishna2017visual}, which allowed it to learn to identify critical regions in the input images that might contain answers to the corresponding questions.

Subsequently, these visual encodings were combined with the sequence of question tokens and then fed into a VisualBERT-based cross-modality encoder, which contains a stack of Transformer layers from a BERT network. These layers effectively leveraged self-attention and cross-attention mechanisms to capture implicit alignments between text phrases and bounding regions. The outputted joint contextual representations of both image and text data was then fed into a shallow classifier, which produced "Yes/No" answers.

\subsection{Proposed Model: Modified VisualBERT with Joint Fusion}

\begin{figure}[h]

\includegraphics[scale=0.15]{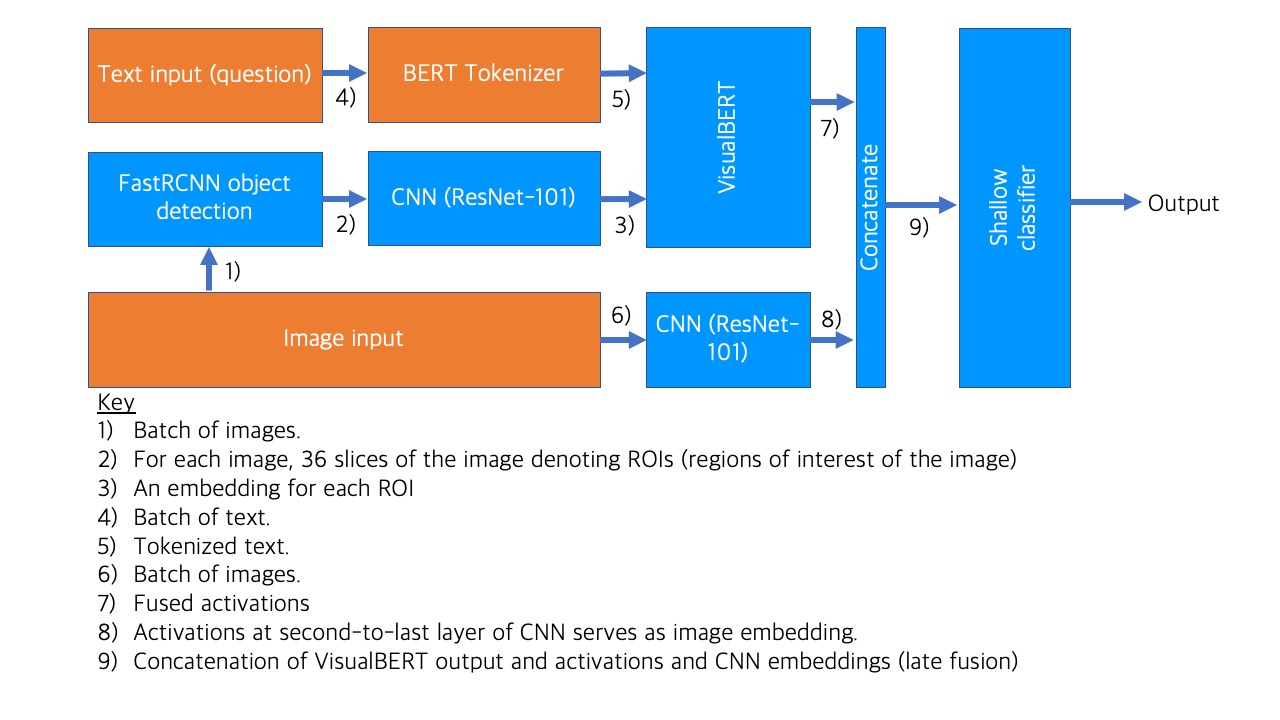}
\caption{Model architecture of the modified VisualBERT with joint fusion}
\label{fig:plot6}
\end{figure}

In addition to the aforementioned model, we propose a second modified VisualBERT model architecture with the addition of joint fusion. In particular, we fed the entire input image into the pre-trained ResNet 101 and obtain a sequence of CNN features. These extracted features were concatenated to the output representations of the cross-modality module and then fed into a shallow classifier, which generate the labels.

The key motivation is that questions on plot data might require understanding the whole images in general instead of analyzing certain regions in the images (such as comparing bar graphs, examining the general trends of line charts, etc.). Thus, we predicted that the CNN features of the entire input images would provide "big picture" information that might help answering the corresponding questions correctly.

\section{Experiments}


As mentioned in the introduction, we designed the experiment in a way to keep the methodology as fair as possible, including keeping the architectures as similar as possible. However, due to the nature of VisualBERT (its input and output), where it requires text tokenized by BERT Tokenizer, there are some unavoidable differences between the models that use VisualBERT and the baselines, that we will highlight here. First, the baseline model uses LSTM in lieu of BERT to process the text, so a tokenizing method with good compatibility with LSTM was used (word tokenizing). Also, GloVe \cite{pennington2014glove} embeddings were used in lieu of BERT embeddings due to the same reason (the word tokenizing instead of sub-word tokenizing). That said, the image preprocessing pipeline, image feature extractors, and output shallow classifiers were exactly the same.

Since our problem can be formulated as a classification task, we used classification accuracy, precision (on the "Yes" label), recall (on the "Yes" label), and F1 score for evaluation to get a comprehensive understanding on the overall performance of the model (accuracy and F1) as well as what kinds of errors the model does (precision and recall).

\section{Results and Analysis} 

\subsection{Model Performance}

The performance of different models on the validation set is shown in Table \ref{results}.

\begin{table*}
\centering
\begin{tabular}{lllll}
\hline
\textbf{Model} & \textbf{Accuracy} & \textbf{Precision} & \textbf{Recall} & \textbf{F1 Score}\\
\hline
Baseline & $0.8366$ & $0.8300$ & $0.8466$ & $0.8382$ \\
Modified VisualBERT & $0.8070$ & $0.7900$ & $0.8200$ & $0.8100$ \\
Modified VisualBERT (Join Fusion) & $0.5815$ & $0.5900$ & $0.5500$ & $0.5700$ \\
\hline
\end{tabular}
\caption{Performance of different models on the validation set.}
\label{results}
\end{table*}

Our implementation of the LSTM + CNN + Shallow Classifier baseline was trained for $8$ epochs and achieved an accuracy of $83.66\%$. The same model with dropout turned off achieved an slightly inferior accuracy of $83.14\%$ at $6$ epochs and started overfitting afterwards. We observe that the baseline can reach a decent level of accuracy with even $1$ epoch of training, with different runs ranging from $82.4\% $to $82.8\%$ accuracy at $1$ epoch.

The two variants of the modified VisualBERT models was trained for $100$ epochs and achieved a decent accuracy of $80.70\%$, along with an F1 score of $0.81$. We can see that the models with the VisualBERT module perform worse than the baseline model which just uses concatenation as fusion, with the baseline's accuracy of $83.66\%$, versus Modified VisualBERT getting an accuracy of $80.70\%$ and Modified VisualBERT with Joint Fusion getting an accuracy of $58.15\%$. This situation is not only true for the Accuracy metric, as the VisualBERT models perform worse than the baseline for Precision, Recall, and F1 Score metrics, with Modified VisualBERT getting $79.00\%$ precision versus the baseline's $83.00\%$ precision, $82.00\%$ recall versus the baseline's $84.66\%$ recall, and $81.00\%$ F1 score versus the baseline's $83.82\%$. This tells us that the VisualBERT approach was able to leverage attention mechanisms to some degree in order to align information in the plots and questions, but it needed a lot of time to learn the relationships in a working level of detail. It also achieves a slightly lower accuracy level than the baseline, which goes against our main hypothesis. This means that for the task of plot question answering, a VisualBERT approach is not recommended. 

At the same time, the addition of joint fusion leads to a significant drop in the accuracy score (by about $20\%$). In other words, adding CNN features of the whole images turns out to hurt the performance of the modified VisualBERT model. An explanation might be that the "big picture" information from the entire input plots actually confuses the classifier, as it does not know which regions to focus on when answering the corresponding question.

\subsection{Error Analysis}

In this section we take a look into the types of questions that each model gets wrong to compare and contrast the different models, highlighting the strengths and weaknesses of the models, and by doing so, the different fusion mechanisms behind the models. We mainly focus on classifying these errors into false positives and false negatives. We also present our guesses on how these tendencies toward false positives and false negatives shed light on the biases, either towards the "Yes" label or the "No", the models exhibit when faced with different examples with distinctive features (including rich logic, dense plot elements, and unseen plot styles).
Hypothesis?

\subsubsection{False positives}

\begin{figure}[h]
\centering
\includegraphics[scale=0.23]{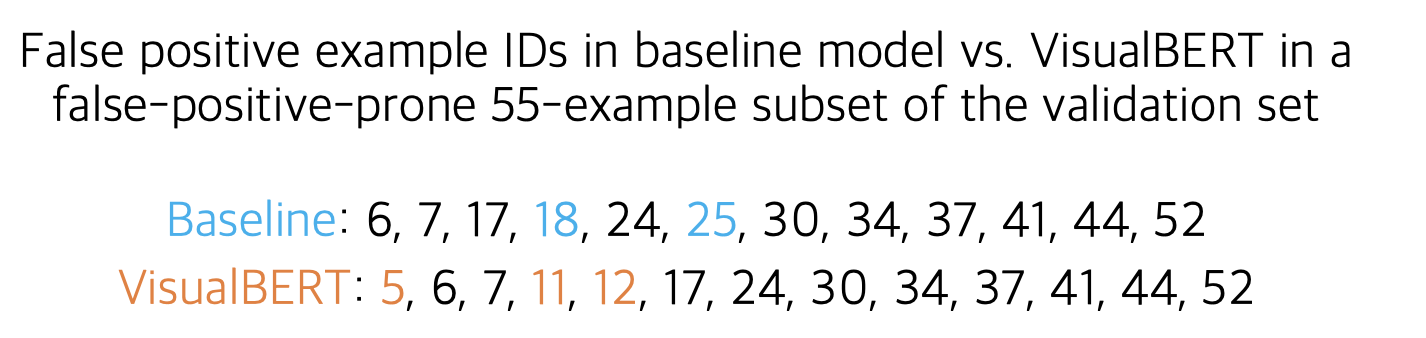}
\caption{False positives comparison between the baseline and modified VisualBERT.}
\label{fig:false_pos}
\end{figure}

First, from Figure \ref{fig:false_pos}, notice that $9$ out of the $11$ false positive errors in baseline and $10$ out of the $13$ false positive errors in Modified VisualBERT were on the same examples. This probably means that those examples were very difficult. For example, some of these examples are bar graphs with a large number of bars, which might cause difficulty for the models when searching for information. It also suggests that there is a chance the models may err towards "Yes" on logically complex questions, although that would require further investigation.

\begin{figure}[h]
\centering
\includegraphics[scale=0.15]{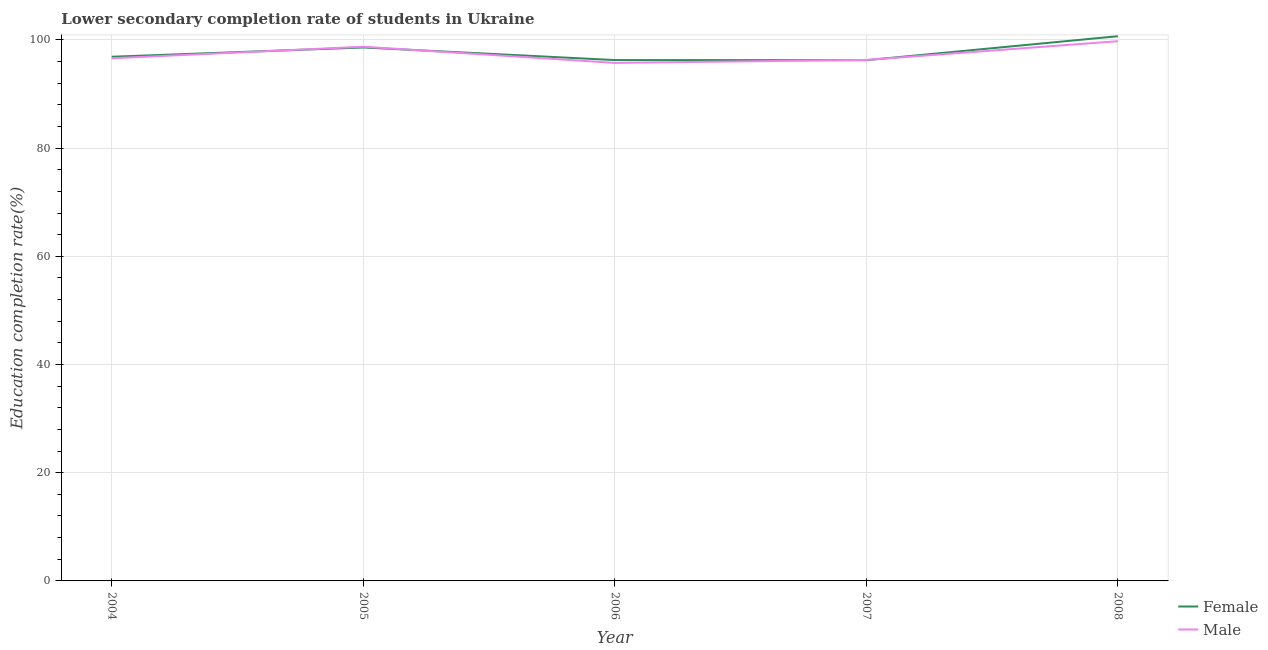}
\caption{Example 25's question is: "Does the education completion rate of female students monotonically increase over the years?" The answer is "No".}
\label{fig:ex25}
\end{figure}

Next, we look at example 25 (Figure \ref{fig:ex25}), which the baseline had an error on but VisualBERT got right. Here, the line only covers a small portion of the image, so VisualBERT with an object detector is able to extract a bounding box that contain the line. Then, it is able to zoom in and closely analyze the trend of that line. On the other hand, the baseline model examines the image as a whole and does not have the ability to detect the small movements of the line, thus it might consider the whole line as having a slightly positive slope. This is an example where object detection in the modified VisualBERT architecture is essential for answering the question.

 
 \subsubsection{False negatives}
 
\begin{figure}[h]
\centering
\includegraphics[scale=0.23]{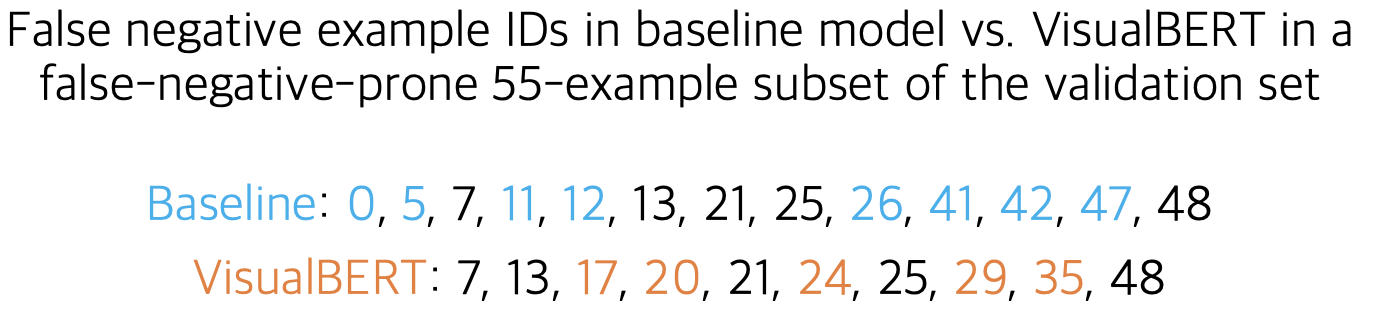}
\caption{False negatives comparison between the baseline and the modified VisualBERT.}
\label{fig:false_neg}
\end{figure}

We notice from Figure \ref{fig:false_neg} that there are fewer false negative errors that the baseline and VisualBERT had in common, with $5$ out of the $13$ false negative errors in baseline and $5$ out of the $10$ false negative errors in VisualBERT being the same examples. This suggests that the examples may not have been too hard logically but may have had diverse formats and many plot elements, where different models may have struggled on different elements. It also suggests that there is a chance that the model errors towards "No" for graphs in a less commonly seen format, although that would need further investigation.

\begin{figure}[h]
\centering
\includegraphics[scale=0.15]{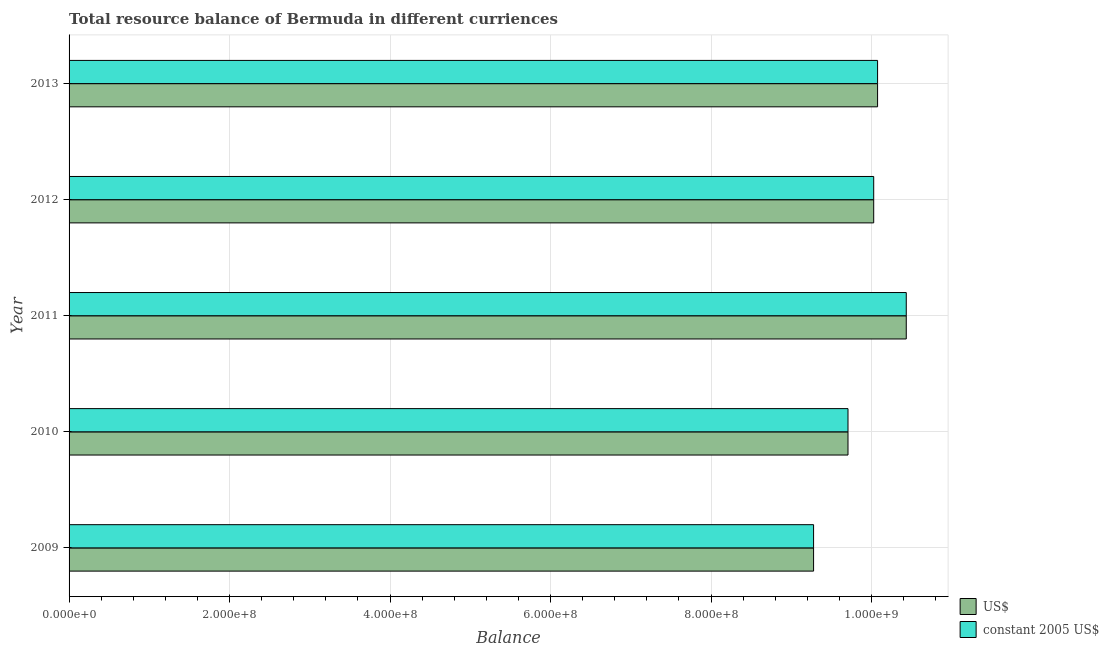}
\caption{Example 35's question is: "Is the resource balance in constant US\$ in 2009 less than that in 2012?" The answer is "Yes".}
\label{fig:ex35}
\end{figure}

Here, we look at example 35 (Figure \ref{fig:ex35}), which is a plot of two columns across five different settings. In this example, there are several columns with similar colors, so after extracting region proposals, the modified VisualBERT model might face difficulty aligning the bounding box with the correct text phrases in the question, namely the years 2009 and 2012. Thus, the model produces an incorrect answer.

\subsubsection{Summary of Error Analysis} 

Overall, the two models were more similar than different in the error analysis on false positives and negatives, which suggests that the vastly different fusion mechanisms may not have a significant impact on this particular aspect of the model performance (including the biases towards "Yes" and "No" and how they treat false negatives or false positives).









\section{Conclusion} 

In summary, our paper aimed to develop deep neural networks called \textbf{BERT-VQA} that tackled the challenge of visual question answering on plots. Initially, we made a hypothesis that the VisualBERT architecture with an attention-based cross-modality module was of great help in improving the model performance. This motivated us to developed a modified version of the VisualBERT architecture with a pretrained ResNet 101 image encoder. We also proposed another variant of the model with joint fusion added. After that, we compared the outcomes of these models with that of a baseline with a LSTM, a CNN, and a shallow classifier combined. However, the final result demonstrated that our original hypothesis was incorrect.

In the future, we might perform further analysis on why the VisualBERT-based model failed to reach the performance of the baseline. In particular, we would like to feature attribution techniques to investigate how the model aligns the bounding regions with text phrases via attention mechanisms and whether these alignments are reasonable. In addition, since our ResNet 101 was pretrained on datasets that contain natural scenes, we want to fine-tune this image encoder on a plot dataset, because this process would allow the model to extract more accurate bounding boxes and produce better visual encodings. We are also interested in exploring model architectures that are able to answer diverse sets of plot-based questions beyond "Yes/No".

\section*{Known Limitations}

The models and findings of this project are strictly applicable only to the domain of answering “Yes/No” questions on plots. The models were trained on a specific subset of the PlotQA dataset with only “Yes/No” questions. Although it is likely that the findings of the project are transferable to other types of questions, including numerical questions (questions where the answer is a floating point number or an integer), contextual questions (questions where the answer is mentioned in the plot), and open questions (questions where the answer does not come directly from the plot and where the model’s reasoning becomes key), it is uncertain if that is the case until further research is done regarding the model’s efficacy with those types of questions. For those who aim to use the model for other types of questions by installing a custom classifier in lieu of our binary classifier, the performance of this setup is unknown and will have to be tested.

For the use as part of a deployed system, we have four major areas of caution. 

\textbf{Domain mismatch}: Aside from the above warning regarding “Yes/No questions only”, please also note that the dataset that we trained the model on, PlotQA, only contain a few types of plots: bar plots, line plots, and dot-line plots. Hence, it is unknown how the model will perform on other plots, including 3-D plots, box plots, stem plots, and graphs, but it is more likely that there will be a drop in performance. Be warned. (This also means that all the findings are specific towards an specific aggregation of bar plots, line plots, and dot-line plots as well.) We recommend training the model using the method described in this paper on your own dataset consisting of the types of plots you would like the model to work on.

\textbf{Other input distribution differences}: Please note that the PlotQA dataset is semi-synthetically generated. All the images are generated programmatically and thus have the same font, look, and feel. This style may be different from the style present in the your application. Real-world plots are also stylistically (and semantically) ever-changing. All of this means that you may need to collect a sample of your desired plots (around 5,000 images is recommended) and run covariate shift correction \cite{covariatecorrection} before you use the model in your deployed system to avoid a ever-increasing decrease in performance. 

\textbf{Fake news}: The model can be used to reason about any plot, such as plots present in literature and in articles. However, the model is trained to reason with the data presented to reach a conclusion, and not to detect the truth or falsehood of the data in the plot. Therefore, any attempt to use the model for such purposes is discouraged. When using the model to automatically analyze plots, it is necessary like always to first verify if those plots are reliable and come from reliable sources.

\textbf{Hardware requirements}: Since we are using BERT, it is a very large model that requires a lot of compute. For inference, a GPU at least on the level of GeForce 1080Ti or Tesla T4 is still recommended, as even a high-quality CPU will take at least 10 - 15 seconds for each image. Since it is out of the scope of this project, we have yet to use network pruning, knowledge distillation, or quantization to lower this requirement slightly.




\bibliography{anthology,custom}

\begin{thebibliography}{16}
\expandafter\ifx\csname natexlab\endcsname\relax\def\natexlab#1{#1}\fi

\bibitem[{Ben-Younes et~al.(2017)Ben-Younes, Cadene, Cord, and
  Thome}]{ben2017mutan}
Hedi Ben-Younes, R{\'e}mi Cadene, Matthieu Cord, and Nicolas Thome. 2017.
\newblock Mutan: Multimodal tucker fusion for visual question answering.
\newblock In \emph{Proceedings of the IEEE international conference on computer
  vision}, pages 2612--2620.

\bibitem[{Chen et~al.(2015)Chen, Fang, Lin, Vedantam, Gupta, Doll{\'a}r, and
  Zitnick}]{chen2015microsoft}
Xinlei Chen, Hao Fang, Tsung-Yi Lin, Ramakrishna Vedantam, Saurabh Gupta, Piotr
  Doll{\'a}r, and C~Lawrence Zitnick. 2015.
\newblock Microsoft coco captions: Data collection and evaluation server.
\newblock \emph{arXiv preprint arXiv:1504.00325}.

\bibitem[{Chen et~al.(2020)Chen, Li, Yu, El~Kholy, Ahmed, Gan, Cheng, and
  Liu}]{chen2020uniter}
Yen-Chun Chen, Linjie Li, Licheng Yu, Ahmed El~Kholy, Faisal Ahmed, Zhe Gan,
  Yu~Cheng, and Jingjing Liu. 2020.
\newblock Uniter: Universal image-text representation learning.
\newblock In \emph{European conference on computer vision}, pages 104--120.
  Springer.

\bibitem[{He et~al.(2016)He, Zhang, Ren, and Sun}]{he2016deep}
Kaiming He, Xiangyu Zhang, Shaoqing Ren, and Jian Sun. 2016.
\newblock Deep residual learning for image recognition.
\newblock In \emph{Proceedings of the IEEE conference on computer vision and
  pattern recognition}, pages 770--778.

\bibitem[{Krishna et~al.(2017)Krishna, Zhu, Groth, Johnson, Hata, Kravitz,
  Chen, Kalantidis, Li, Shamma et~al.}]{krishna2017visual}
Ranjay Krishna, Yuke Zhu, Oliver Groth, Justin Johnson, Kenji Hata, Joshua
  Kravitz, Stephanie Chen, Yannis Kalantidis, Li-Jia Li, David~A Shamma, et~al.
  2017.
\newblock Visual genome: Connecting language and vision using crowdsourced
  dense image annotations.
\newblock \emph{International journal of computer vision}, 123(1):32--73.

\bibitem[{Li et~al.(2019{\natexlab{a}})Li, Gan, Cheng, and
  Liu}]{li2019relation}
Linjie Li, Zhe Gan, Yu~Cheng, and Jingjing Liu. 2019{\natexlab{a}}.
\newblock Relation-aware graph attention network for visual question answering.
\newblock In \emph{Proceedings of the IEEE/CVF international conference on
  computer vision}, pages 10313--10322.

\bibitem[{Li et~al.(2019{\natexlab{b}})Li, Yatskar, Yin, Hsieh, and
  Chang}]{li2019visualbert}
Liunian~Harold Li, Mark Yatskar, Da~Yin, Cho-Jui Hsieh, and Kai-Wei Chang.
  2019{\natexlab{b}}.
\newblock Visualbert: A simple and performant baseline for vision and language.
\newblock \emph{arXiv preprint arXiv:1908.03557}.

\bibitem[{Lin et~al.(2014)Lin, Maire, Belongie, Hays, Perona, Ramanan,
  Doll{\'a}r, and Zitnick}]{lin2014microsoft}
Tsung-Yi Lin, Michael Maire, Serge Belongie, James Hays, Pietro Perona, Deva
  Ramanan, Piotr Doll{\'a}r, and C~Lawrence Zitnick. 2014.
\newblock Microsoft coco: Common objects in context.
\newblock In \emph{European conference on computer vision}, pages 740--755.
  Springer.

\bibitem[{Methani et~al.(2020)Methani, Ganguly, Khapra, and Kumar}]{plotqa}
Nitesh Methani, Pritha Ganguly, Mitesh~M Khapra, and Pratyush Kumar. 2020.
\newblock Plotqa: Reasoning over scientific plots.
\newblock In \emph{Proceedings of the IEEE/CVF Winter Conference on
  Applications of Computer Vision}, pages 1527--1536.

\bibitem[{Narasimhan et~al.(2018)Narasimhan, Lazebnik, and
  Schwing}]{narasimhan2018out}
Medhini Narasimhan, Svetlana Lazebnik, and Alexander Schwing. 2018.
\newblock Out of the box: Reasoning with graph convolution nets for factual
  visual question answering.
\newblock \emph{Advances in neural information processing systems}, 31.

\bibitem[{Pennington et~al.(2014)Pennington, Socher, and
  Manning}]{pennington2014glove}
Jeffrey Pennington, Richard Socher, and Christopher~D. Manning. 2014.
\newblock \href {http://www.aclweb.org/anthology/D14-1162} {Glove: Global
  vectors for word representation}.
\newblock In \emph{Empirical Methods in Natural Language Processing (EMNLP)},
  pages 1532--1543.

\bibitem[{Reddi et~al.(2015)Reddi, Poczos, and Smola}]{covariatecorrection}
Sashank Reddi, Barnabas Poczos, and Alex Smola. 2015.
\newblock \href {https://doi.org/10.1609/aaai.v29i1.9576} {Doubly robust
  covariate shift correction}.
\newblock \emph{Proceedings of the AAAI Conference on Artificial Intelligence},
  29(1).

\bibitem[{Ren et~al.(2015)Ren, He, Girshick, and Sun}]{ren2015faster}
Shaoqing Ren, Kaiming He, Ross Girshick, and Jian Sun. 2015.
\newblock Faster r-cnn: Towards real-time object detection with region proposal
  networks.
\newblock \emph{Advances in neural information processing systems}, 28.

\bibitem[{Tan and Bansal(2019)}]{tan2019lxmert}
Hao Tan and Mohit Bansal. 2019.
\newblock Lxmert: Learning cross-modality encoder representations from
  transformers.
\newblock \emph{arXiv preprint arXiv:1908.07490}.

\bibitem[{Teney et~al.(2018)Teney, Anderson, He, and Van
  Den~Hengel}]{teney2018tips}
Damien Teney, Peter Anderson, Xiaodong He, and Anton Van Den~Hengel. 2018.
\newblock Tips and tricks for visual question answering: Learnings from the
  2017 challenge.
\newblock In \emph{Proceedings of the IEEE conference on computer vision and
  pattern recognition}, pages 4223--4232.

\bibitem[{Teney et~al.(2017)Teney, Liu, and van Den~Hengel}]{teney2017graph}
Damien Teney, Lingqiao Liu, and Anton van Den~Hengel. 2017.
\newblock Graph-structured representations for visual question answering.
\newblock In \emph{Proceedings of the IEEE conference on computer vision and
  pattern recognition}, pages 1--9.

\end{thebibliography}




\end{document}